%% file: main.tex
\documentclass[runningheads]{llncs}
\usepackage{graphicx}
\usepackage{makeidx}
\usepackage{url}
\usepackage{multirow}
\usepackage{cite}
\usepackage{graphicx}
\graphicspath{{figs/}}
\usepackage{amsmath,bm}
\usepackage{amssymb}
\usepackage{algorithm}
\usepackage{algorithmic}
\begin{document}
\title{Patch-CNN: Training data-efficient deep learning for high-fidelity diffusion tensor estimation from minimal diffusion protocols}
\titlerunning{Patch-CNN: Training data-efficient DL for diffusion tensor estimation}
%
\author{Tobias Goodwin-Allcock\inst{1} \and
Ting Gong\inst{1} \and
Robert Gray\inst{2} \and
Parashkev Nachev\inst{2} \and
Hui Zhang\inst{1}}
\authorrunning{T. Goodwin-Allcock et al.}
%
\institute{Department of Computer Science and Centre for Medical Image Computing, University College London, London, United Kingdom \and Department of Brain Repair \& Rehabilitation, Institute of Neurology, UCL, London, United Kingdom.}
\maketitle              
\begin{abstract}
We propose a new method, Patch-CNN, for diffusion tensor (DT) estimation from only six-direction  diffusion weighted images (DWI). Deep learning-based  methods have been recently proposed for dMRI parameter estimation, using either voxel-wise fully-connected neural networks (FCN) or image-wise convolutional neural networks (CNN). In the acute clinical context—where pressure of time limits the number of imaged directions to a minimum—existing approaches either require an infeasible number of training images volumes (image-wise CNNs), or do not estimate the fibre orientations (voxel-wise FCNs) required for tractogram estimation. To overcome these limitations, we propose Patch-CNN, a neural network with a minimal (non-voxel-wise) convolutional kernel (3×3×3). Compared with voxel-wise FCNs, this has the advantage of allowing the network to leverage local anatomical information. Compared with image-wise CNNs, the minimal kernel vastly reduces training data demand. Evaluated against both conventional model fitting and a voxel-wise FCN, Patch-CNN, trained with a single subject is shown to improve the estimation of both scalar dMRI parameters and fibre orientation from six-direction DWIs. The improved fibre orientation estimation is shown to produce improved tractogram.

\keywords{Patch CNN  \and Machine learning \and Diffusion MRI.}

\end{abstract}

\input{00introduction}

\input{01method}
\input{02results}
\input{03discussion}
\input{04conclusion}
\input{05acknowledgement}

\bibliographystyle{splncs04}
\bibliography{references}

\end{document}

%% file: 00introduction.tex
\section{Introduction}
\label{sec:introduction}

Diffusion tensor (DT) magnetic resonance imaging (MRI) allows for non-invasive quantification of white matter (WM) microstructure and connectivity, providing measures shown to be sensitive to disease~\cite{Bodini2014DiffusionDisorders}.
However, accurate estimation of tissue microstructure---under conventional estimation routines---requires at least 30 diffusion weighted images (DWI)~\cite{Jones2004NumDWIs}. This requirement is incompatible with current acute clinical practice, where commonly only six-direction DWIs can be acquired within the feasible envelope of available time and patient tolerability.

Recently, deep learning (DL) has been proposed to reduce the number of DWIs required for accurate estimation of tissue microstructure parameters~\cite{Golkov2016QSpaceDL, Aliotta2019MLP, Aliotta2020U-Net, Li2019HCNN, Lin2019fODF,Tian2020DeepDTI:Learning}. 
Early works aimed to directly estimate scalar parameters from a reduced number of DWIs and treated each voxel independently. This was first achieved in q-Space deep learning~\cite{Golkov2016QSpaceDL}; this method estimated scalar diffusion kurtosis imaging (DKI)~\cite{Jensen2005DiffusionalImaging} and neurite orientation dispersion and density imaging (NODDI)~\cite{Zhang2012NODDI} parameters---such as radial kurtosis and neurite orientation dispersion index---accurately with 12$\times$ acceleration. This method was then adapted to DiffNet~\cite{Aliotta2019MLP} which accurately estimates scalar diffusion tensor (DT) measures, fractional anisotropy (FA) and mean diffusivity (MD), directly from as little as three DWIs. 
However, none of these methods estimate the primary diffusion direction required for tractography. 

Direction has previously been accurately estimated from six-directional DWIs. This was achieved by DeepDTI~\cite{Tian2020DeepDTI:Learning}, but, this network uses a more complex---image-wise---convolutional neural network (CNN) architecture. This architecture requires a training dataset that contains a representative distribution of the macro-scale brain structure, therefore an order of magnitude more training subjects is required. Additionally, these methods may not be robust to abnormal brain structure e.g. pathology, and may not easily learn pathology given the wide space of possible pathological characteristics. 

Here we aim to estimate high-fidelity DTs from only six DWIs, like DeepDTI, but having a minimal training-data requirement, like DiffNet. To achieve that, we adopt a patch-wise CNN architecture. This approach has been shown to significantly improve estimation accuracy of scalar measures from DT and DKI models compared to FCNs when only a small number of DWIs are available~\cite{Li2019HCNN}. This shows the benefit of incorporating local anatomical information in a judicious manner. Additionally, patch-wise CNNs have been shown to improve fibre orientation estimation~\cite{Lin2019fODF}. In this work, the full fibre orientation distributions were estimated accurately from a patch-wise input, however, this input used a large number ($>$6) of diffusion-encoding directions. We therefore propose to adopt a patch-wise architecture to estimate DTs with only six-directional DWIs as input.

We organise the rest of the paper as follows: Section \ref{sec:method} describes the Patch-CNN method and how we assess it; Section \ref{sec:results} shows the results which are then discussed in Section \ref{sec:discussion}.

%% file: 01method.tex
\section{Methods}
\label{sec:method}
\subsection{Proposed deep learning technique}
Here, we propose Patch-CNN, an adaptation H-CNN~\cite{Li2019HCNN}, for DT estimation from only six DWIs. 
To minimise training data requirements, we must \textit{avoid} learning macro-scale brain anatomy, for an inadequately learnt global pattern may be misleading. Therefore, the network's input size needs to be minimal.
The minimum input size is a voxel-wise network; however, we will show later that voxel-wise networks do not improve estimation for the directional parameters.
Increasing the input window to a minimal (non-voxel-wise) 3D patch (3$\times$3$\times$3) has been shown to increase estimation accuracy~\cite{Li2019HCNN} (H-CNN) for scalar parameter estimation. This is attributed to combining local neighbourhood information.
Additionally, H-CNN does not rely upon macro-scale brain anatomy.
The proposed method, Patch-CNN, is based on the previous methods architecture except it is adapted for estimation of the DT by estimating the six values of the DT.
As the hierarchical aspect is no longer required we estimate all six values from the final layer of the network.
To ensure semi-positive definiteness of the estimated DT's, the network is trained to estimate the matrix log of the diffusion tensor.

In summary, Patch-CNN consists of a convolutional layer with a 3x3x3 kernal and a ReLu activation function followed by 2 fully-connected hidden layers each with 150 units and ReLu activation functions finally outputting to the 6 independent parameters of the matrix log of the diffusion tensor.

\subsection{Evaluation method}
\subsubsection{Baseline techniques}
To assess the proposed method we compare against conventional model fitting (Conventional) and voxel-wise machine learning (Voxel-NN) at DT estimation from six DWIs.  
Voxel-NN uses the same architecture as Patch-CNN except for the kernel size of the input layer which is reduced to 1$\times$1$\times$1. This is equivalent to a voxel-wise fully-connected layer of 150 units.
This architecture is similar to other voxel-wise networks~\cite{Golkov2016QSpaceDL, Aliotta2019MLP}.
For comparison against conventional model fitting we apply linear least squares using FSL~\cite{Smith2004AdvancesFSL}.

\subsubsection{Tractography}
For all methods, tractograms are derived from estimated DTs using the FACT algorithm~\cite{Mori1999FACT}. The FACT algorithm generates streamlines in an iterative process starting at seed points in the white matter (WM) and follows the primary fibre orientation until it terminates; we used the FACT algorithm implemented in DTITK~\cite{DTITK}. WM masks are defined as being voxels with estimated linearity~\cite{Westin2002Linearity} $>$ 0.6. The tractogram is separated into tract bundles by including all streamlines which pass through specific regions of interest (ROI)~\cite{wakana2004TractBundling1, Stieltjes2001TractBundling2}. The corticospinal tract (CST) and corpus callosum (CC) bundles are chosen for assessment as they are well known large structures whose fibres are tangential to each other.

\subsubsection{Dataset}
For this evaluation we require a dataset for training and testing; this dataset must contain a large set of DT-compatible measurements to provide 1) a high-quality ‘ground-truth’ (GT) DT and tractogram using conventional fitting 2) DWI subsets with fewer measurements that mimic six directional DWI gradient schemes.
These specifications are satisfied by the Human Connectome Project (HCP)~\cite{HCP} dataset due to it's DTI-compatible measurements that include b=0’s (18) and b=1000’s (90). All of these DWIs are used to estimate the GT diffusion tensors; estimation is performed using conventional linear least squares fitting.
To imitate a clinical scan consisting of a single b=0 image and an optimal subset of six b=1000’s we chose to subsample the gradient scheme to be maximally similar to Skare’s~\cite{Skare2000SampScheme} six DWI gradient scheme. This scheme minimises the condition number of the bvec matrix and is used in DeepDTI~\cite{Tian2020DeepDTI:Learning}.

\begin{figure}[hb!]
\begin{center}
\includegraphics[width=\linewidth] {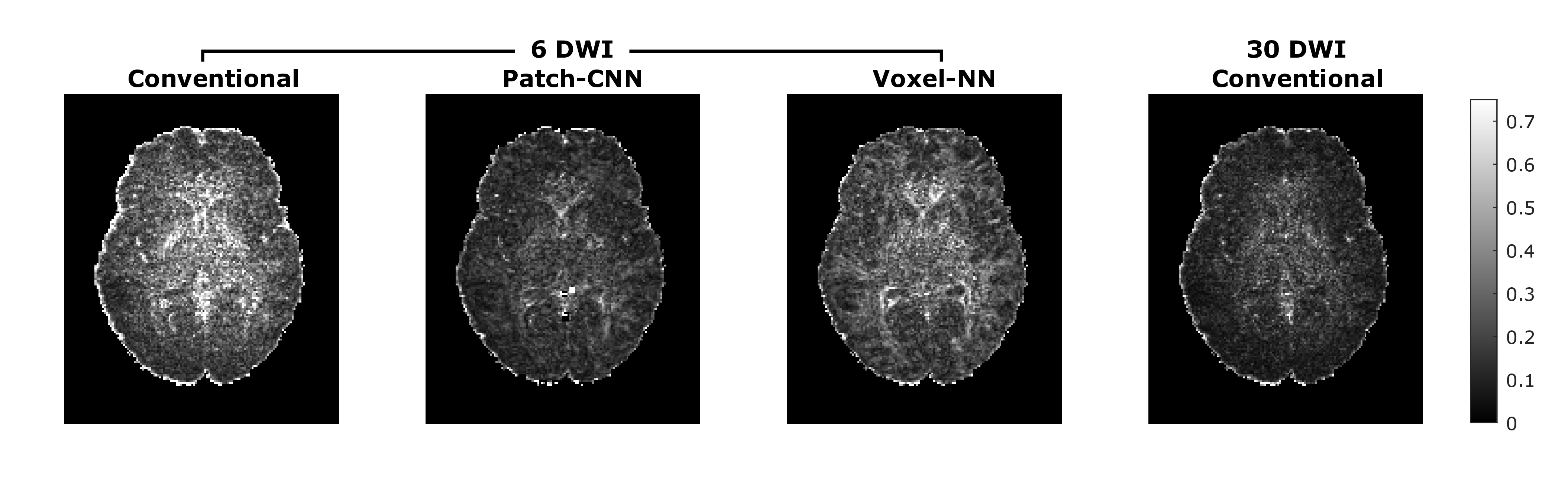}
\caption{
The Frobenius norm of the error map between the estimations and ground truth image. Patch-CNN has the lowest error of any of the estimations from six DWI. The error is slightly higher than the error map of conventional-fitting applied to 30 DWIs.
}
\label{fig:frobNormImg}
\end{center}
\end{figure}

\subsubsection{Training}
Both the proposed and baseline ML method are trained using the same training settings.
To mimic minimal training data requirements, data from a single HCP subject was used; this data consists of the optimal six DWI subset, for the input, and the matrix-logarithm of the ‘ground-truth’ DTs, for the output.
Training settings follow the implementation of H-CNN: a batch size of 256; optimisation uses the ADAM optimiser~\cite{Kingma2015Adam:Optimization}; the learning rate is set to 0.001 initially and subsequently reduced by 50\% at each plateau. 
Plateaus in training are defined by no improvement in training loss over 10 consecutive epochs. 
To reduce overfitting, early stopping is applied with a random choice of 20\% of the brain voxels as validation, with the remaining 80\% for training.

\subsubsection{Testing}
To test all of the methods we estimate DTI’s for 12 subjects---unseen during training---using each method and measure the error. 

\subsection{Evaluation Metrics}
Estimation error is computed on the estimated DT’s as well as the derived scalar parameters, directional parameters and tractograms.
To assess the ability to estimate the tensors and scalar parameters, the errors between estimated and GT measurements are calculated over the whole brain. The metrics used are the Frobenius norm for tensors and absolute difference for FA and MD. 
To assess its ability to estimate the directional measures---required to reconstruct WM bundles---the primary fibre orientation estimation error is calculated. 

For the primary fibre orientation the angular errors between the estimated primary fibre orientation and the ones from GT tensors are computed over the WM, where white matter is determined by voxels with a ground truth linearity coefficient~\cite{Westin2002Linearity} $>$ 0.6. 

The accuracy of generated tract bundles is assessed on a tract bundle level. 
For each bundle the accuracy is calculated by the similarity between the estimated tract bundles and the ground truth tract bundle. This is computed by the dice overlap of the voxels containing at least 25 streamlines.

\begin{figure}[ht!]
\begin{center}
\includegraphics[width=\linewidth] {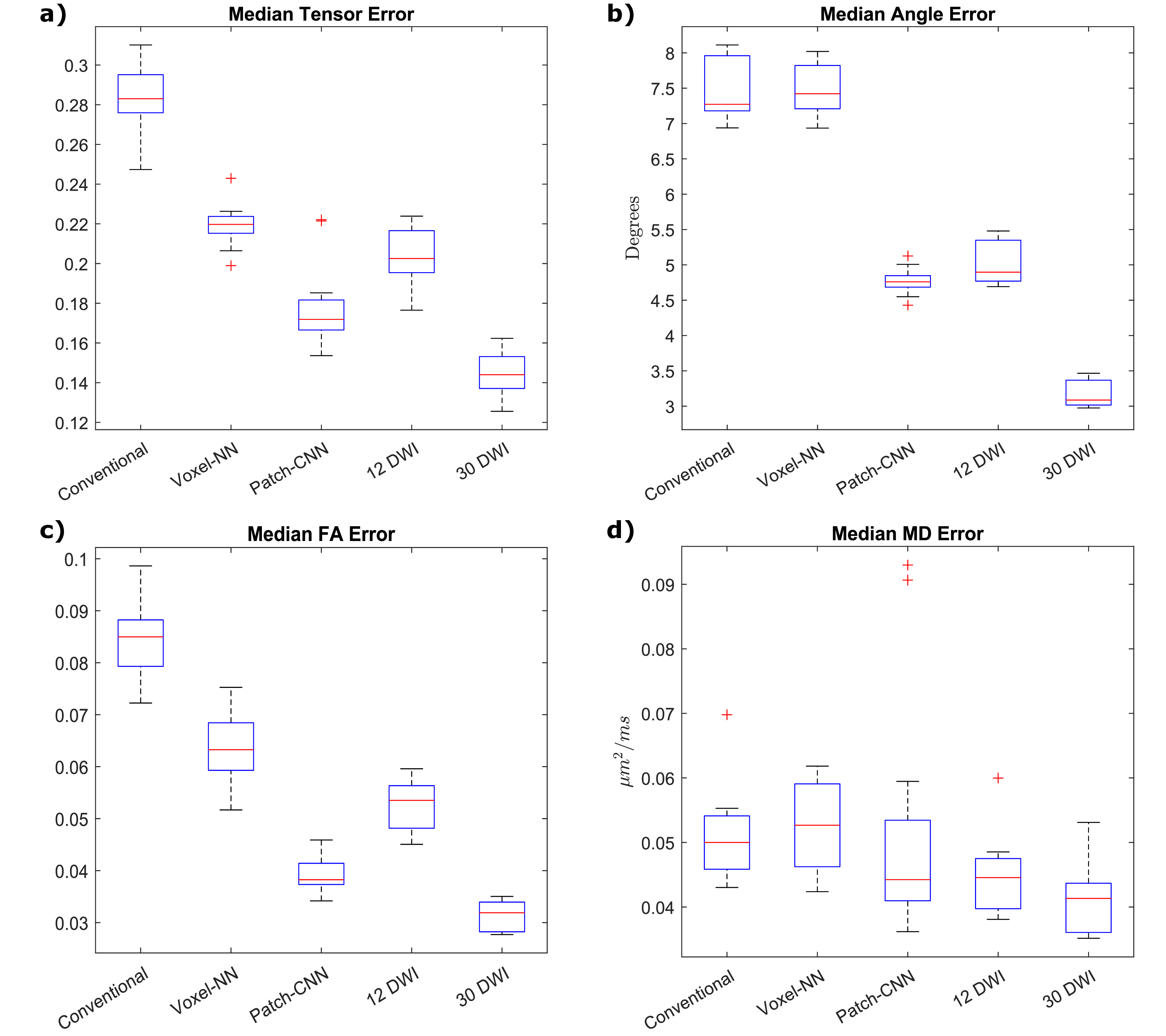}
\caption{
Boxplots computed over the medians of errors for each of the 12 testing subjects. For (a,c,d) the median error is computed across all of the brain voxels at each subject. For (b) median error is computed for each subject across voxels for which the primary direction of diffusion is well defined, where the linearity coefficient~\cite{Westin2002Linearity} $>$0.6. Patch-CNN's estimates with higher fidelity than Voxel-NN and methods  conventional fitting with double the number of directions.
}
\label{fig:boxplots}
\end{center}
\end{figure}

%% file: 02results.tex
\section{Results}
\label{sec:results}

Figure \ref{fig:frobNormImg} shows the error of the estimated tensors---measured by the Frobenius norm---over an example axial slice for each of the different estimation techniques.
On six directional data, the estimation technique with the lowest error is Patch-CNN; this method performs better than both Voxel-NN and conventional fitting.
Additionally, the error map for Patch-CNN is comparable with 30 directional conventional fitting.
These qualitative results are affirmed by the boxplots in Figure \ref{fig:boxplots}a.
Here, each boxplot corresponds to median of the Frobenius norm for each of the 12 subjects for each of the methods. 
Not only is Patch-CNN shown to outperform all methods estimating from the same number of directions, it also outperforms conventional fitting from twice as many DWIs.

\begin{figure}[ht!]
\begin{center}
\includegraphics[width=\linewidth] {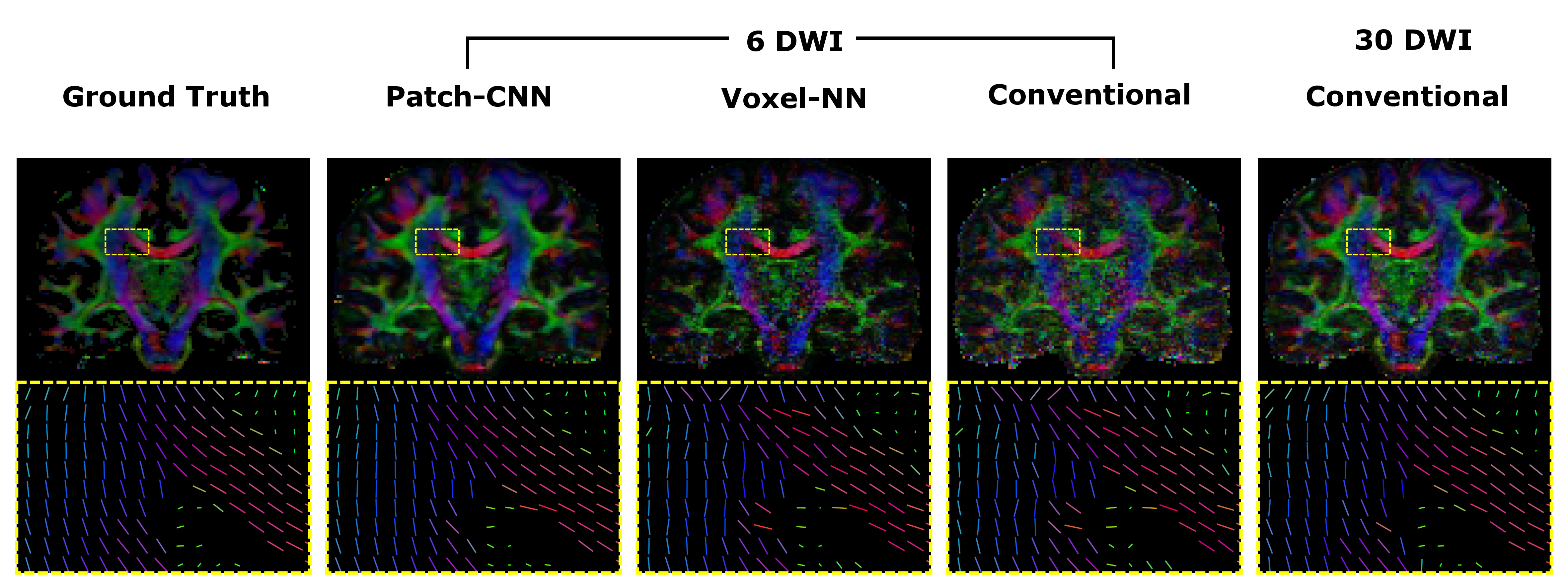}
\caption{
Top are the FA weighted colour map of the primary direction of diffusion. Part of the motor tract and corpus callosum, highlighted in yellow, is enlarged at the bottom where the primary directions of diffusion are illustrated as colour encoded sticks. The sticks are masked such that only the WM voxels remain, determined by FA$>$0.2. Estimations from Patch-CNN are visually more similar to the GT for both RGB colourmap and sticks.
}
\label{fig:V1}
\end{center}
\end{figure}

\begin{figure}[ht!]
\begin{center}
\includegraphics[width=\linewidth] {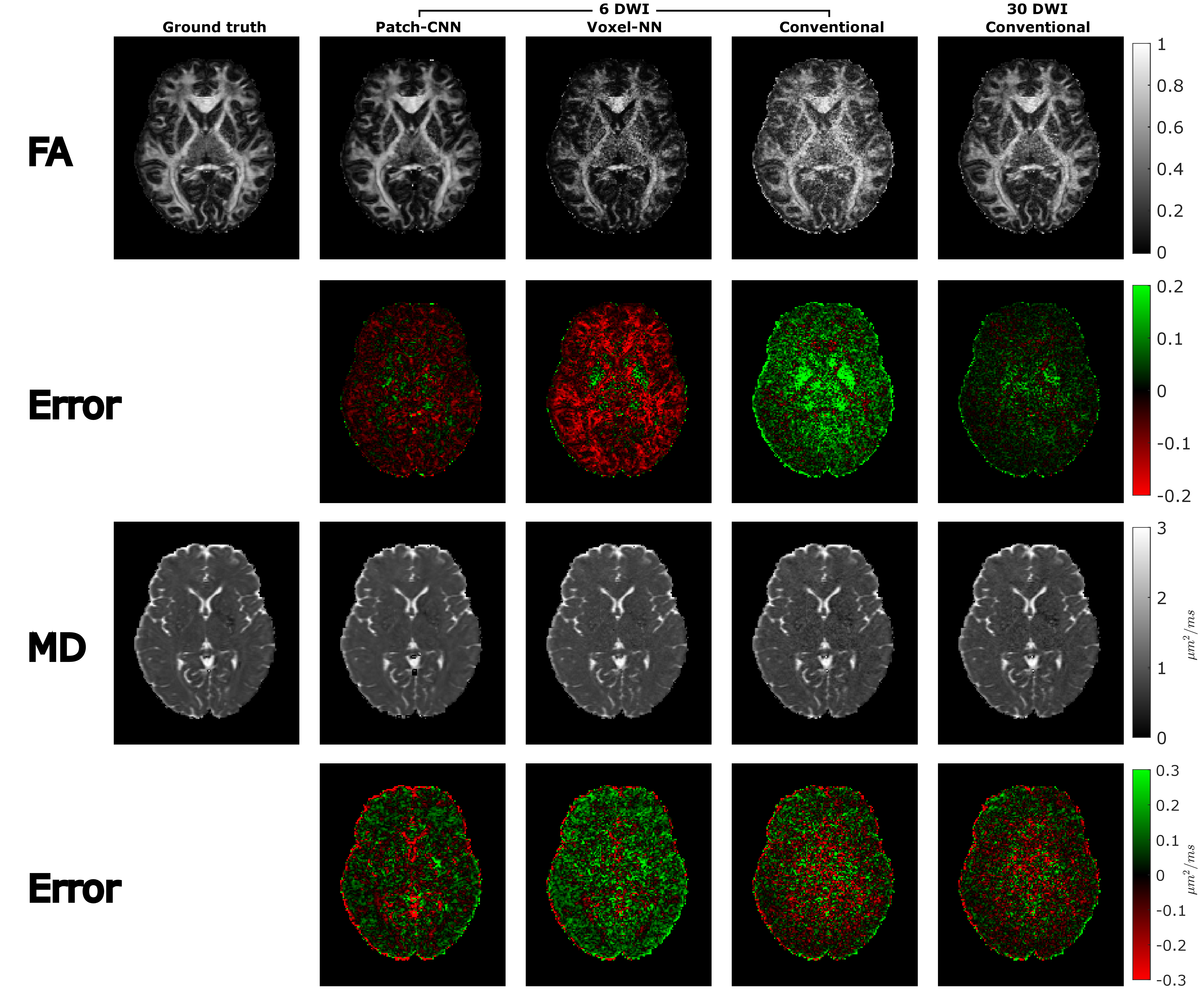}
\caption{
Scalar parameter (FA and MD) estimation are shown with their corresponding difference maps. Patch-CNN estimates less grainy and more accurate images for both MD and FA when comparing against the model fit and Voxel-NN. The improvement is larger in FA estimation.
}
\label{fig:FAMD_diff}
\end{center}
\end{figure}

\begin{figure}[ht!]
\begin{center}
\includegraphics[width=\linewidth] {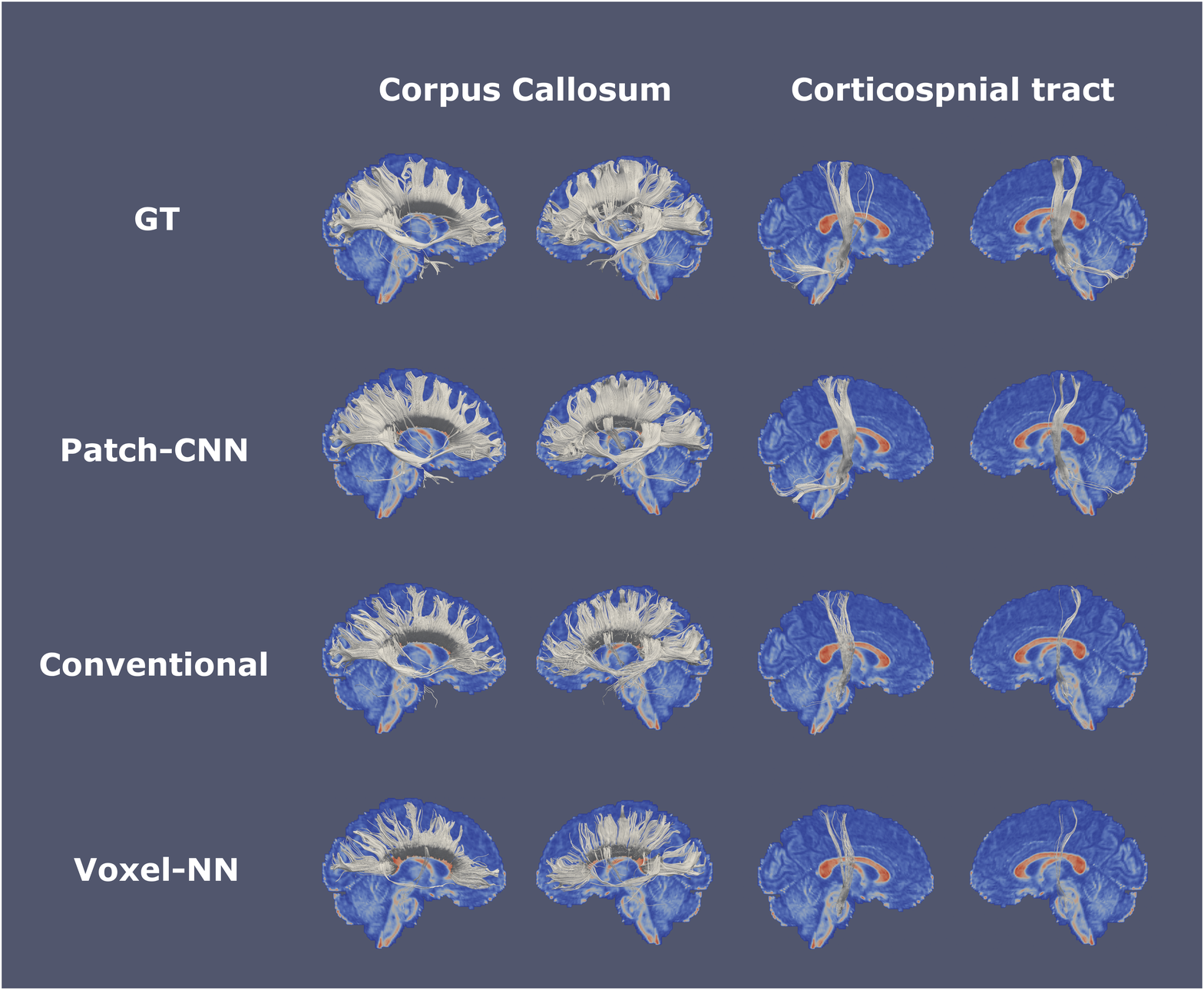}
\caption{
Estimations for the corpus callosum and corticospinal tract achieved by performing the FACT algorithm on ground truth tensors as well as tensors estimated from six DWIs using Patch-CNN, conventional and Voxel-NN fitting. A mid sagittal slice of the ground truth FA is used to show the relative locations of the tracts to local white matter as well as separating the left and right tract bundles. Patch-CNN performs the most robust tractogram estimation of all the methods as the tract bundles are far denser.
}
\label{fig:tract}
\end{center}
\end{figure}

\begin{figure}[ht!]
\begin{center}
\includegraphics[width=\linewidth] {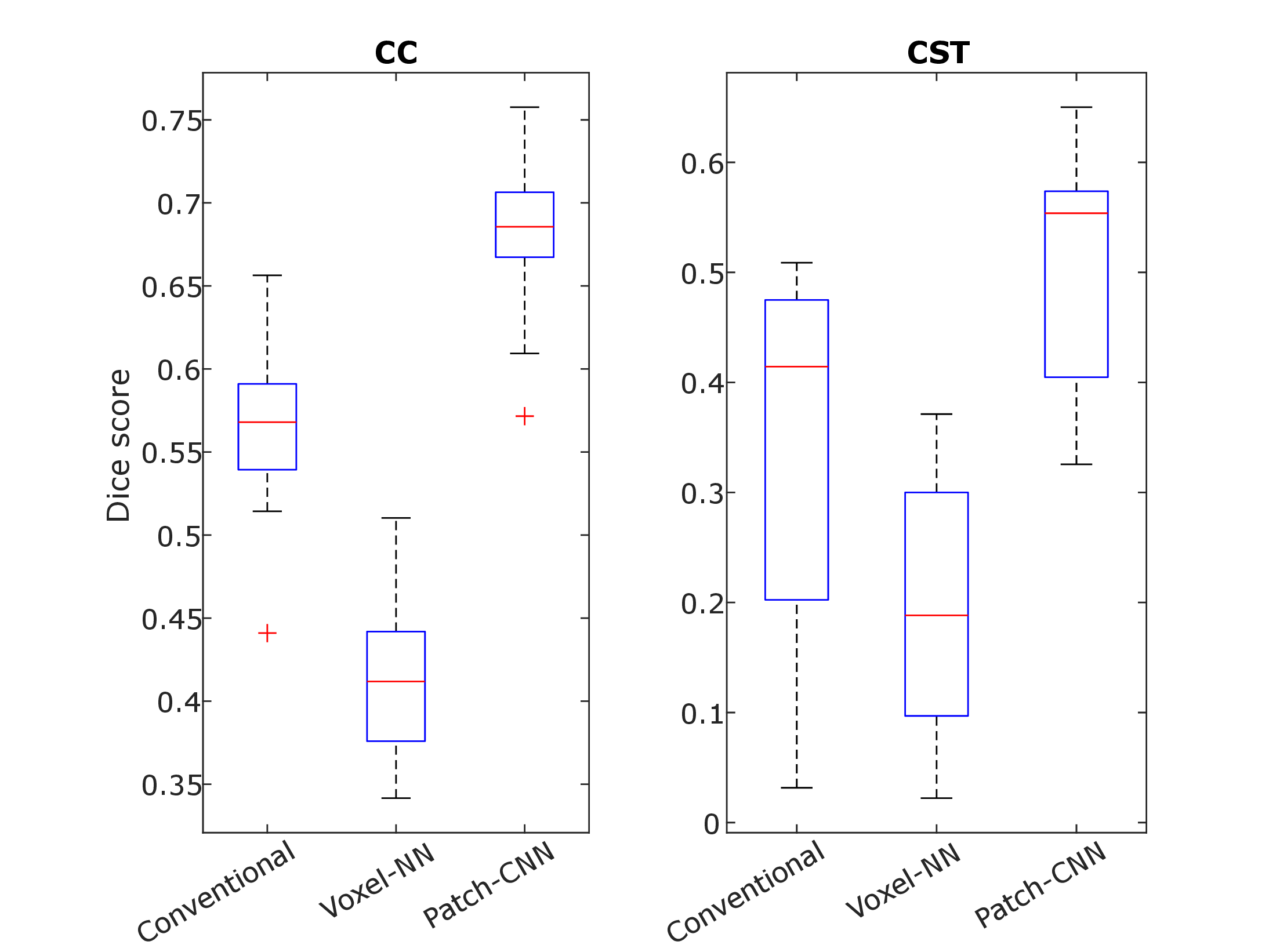}
\caption{
An assessment of the similarity between the ground truth tractogram and the tractogram computed from diffusion tensors estimated using the three methods for one subject. The similarity metric compared binary images derived from the streamlines by assigning one to a voxel if over 25 streamlines pass through it. The similarity between GT and estimated binary images is assessed using dice overlap. The dice overlap across the 12 testing subjects is shown in the boxplots. Patch-CNN's estimates are the most similar to the ground truth as the dice overlap is higher than both of the other methods.
}
\label{fig:boxplotOfTract}
\end{center}
\end{figure}

Figure \ref{fig:V1} shows a coronal slice of the FA weighted colour encoded primary fibre orientation for the ground truth and estimations across multiple fitting methods. Qualitatively, the Patch-CNN estimate is far more accurate than both of its six directional counterparts as it is both more visually similar to the ground truth and less noisy. To see the effect in more detail we enlarge a small region of the motor tract and corpus callosum, shown in the yellow boxes. In this enlarged view the primary fibre orientations, visualised as sticks, estimated by Patch-CNN are coherent like the ground truth and 30 directional conventional estimation. In contrast both of the other estimations from six directions have very incoherent fibres. Figure \ref{fig:boxplots}b confirms this quantitatively over the 12 testing subjects and again shows that Patch-CNN outperforms conventional fitting with twice as many DWIs.

The estimation performance on scalar measures derived from the diffusion tensor is shown in Figure \ref{fig:FAMD_diff}. Axial slices of the estimated FA and MD are shown above their error maps. Qualitatively, Patch-CNN performs the best of the six directional estimations because it is the most similar to the ground truth image as well as having the least amount of noise. For FA estimation, conventional fitting largely overestimates the FA whereas Voxel-NN largely underestimates the FA. Only Patch-CNN estimates FA with far greater accuracy. Figures \ref{fig:boxplots}c and \ref{fig:boxplots}d further affirms that Patch-CNN outperforms conventional fitting with twice as many DWIs.

Figure \ref{fig:tract}, shows the tract bundles of the corpus callosum and corticospinal tract. Qualitative assessment shows that Patch-CNN performed the best, the CC bundle is far denser and the left CST has streamlines terminating in two regions just like the ground truth. Figure \ref{fig:boxplotOfTract} quantitatively supports this as Patch-CNN has the highest average dice score of the three methods on all of the tract bundles.

%% file: 03discussion.tex
\section{Discussion}
\label{sec:discussion}
We have developed a data-efficient method of estimating the diffusion tensor with high accuracy, leading to improved performance in all derived parameters both scalar and directional, as well as greatly improved estimates of the tractograms, in comparison with rival methods. Patch-CNN outperformed conventional fitting as well as Voxel-NN at estimating the corpus callosum and corticospinal tract. The most improvement was shown in the estimation of the corticospinal tract. Although the corticospinal tract was the tract estimation most improved by Patch-CNN it was still also worse than the CC estimation. We believe that both of these factors are due to the length of the corticospinal streamlines being longer than that of the CC streamlines; longer tracts travel through a larger number of voxels and so premature termination is much more likely.

Tensors estimated from Voxel-NN have a lower average Frobenius norm on the error compared to conventional fitting (0.22, 0.28) this is reflected in the accuracy of the scalar measures MD and FA and is consistent with the literature. However, Voxel-NN's estimation of the primary fibre orientation is not as accurate as conventional fitting. It is no surprise that the tractograms---determined by the primary fibre orientation---also shows a decrease in performance when using Voxel-NN. Therefore, we see that the increase in estimation accuracy for primary fibre orientations and the resultant tractograms from Patch-CNN is due to including local neighbourhood information into the input.

Although we have only shown the accuracy of MD and FA scalar measures, as the whole diffusion tensor is estimated, any scalar measure which can be derived from the diffusion tensor---such as radial diffusivity, linearity, planarity and sphericality---can be estimated with no new training regime required.

Patch-CNN is more clinically viable than image-wise ML models due to Patch-CNN’s modest training data requirements. Patch-CNN only requires a single subject for training because it’s small input field size means that each subject’s data contains hundreds of thousands of training examples. For an ML model whose input field size is the entire image, a single subject’s data is a single training example. Therefore, a much larger number of subjects must be recruited to train image-wise models. This is especially important as each model is tied to the acquisition settings, e.g. gradient scheme, of the data it was trained on. For every new set of acquisition settings more training data must be collected. As Patch-CNN only requires one subject to train, it is still clinically feasible.

A limitation of the proposed method, along with Voxel-NN, is a tendency to underestimate the high FA regions. A possible cause of this is class imbalance as high FA voxels are vastly underrepresented in the training data. Class imbalance could be combated with data balancing of the training data.

Robustness to lesion presence was not shown and is left as future work.

%% file: 04conclusion.tex
\section{Conclusion}
\label{sec:conclusion}
Patch-CNN has been demonstrated to outperform both conventional fitting and voxel-wise machine learning for estimating the DT and the clinically useful measures derived from it, matching the performance of conventional fitting with twice the number of DWIs. Tractograms derived from Patch-CNN’s tensor estimates have higher fidelity than from conventionally estimated tensors. Patch-CNN achieves this performance with only one training subject which is a vast improvement from the image-wise CNNs such as DeepDTI which uses around 40 training subjects as well as both T1w and T2w anatomical images. Patch-CNN allows for robust estimation of tracts from only six DWI's in a clinically feasible manner.

%% file: 05acknowledgement.tex
\section*{Acknowledgements}
\label{ack}
This work is supported by the EPSRC-funded UCL Centre for Doctoral Training
in Medical Imaging (EP/L016478/1), the Department of Health’s
NIHR-funded Biomedical Research Centre at UCLH and the Wellcome Trust.